# Natural Disaster Analysis using Satellite Imagery and Social-Media Data for Emergency Response Situations

Sukeerthi Mandyam*[1], Shanmuga Priya M.G[2]., Shalini S[3]., Kavitha S[4]

[1, 2, 3, 4] Dept of Computer Science and Engineering,

[1, 2, 3, 4] Sri Sivasubramaniya Nadar College of Engineering, Chennai, Tamil Nadu, India

*Corresponding author:

E-mail: sukeerthi18175@cse.ssn.edu.in

# ABSTRACT

Disaster Management is one of the most promising research areas because of its significant economic, environmental and social repercussions. This research focuses on analyzing different types of data (pre and post satellite images and twitter data) related to disaster management for in-depth analysis of location-wise emergency requirements. This research has been divided into two stages, namely, satellite image analysis and twitter data analysis followed by integration using location. The first stage involves pre and post disaster satellite image analysis of the location using multi-class land cover segmentation technique based on U-Net architecture. The second stage focuses on mapping the region with essential information about the disaster situation and immediate requirements for relief operations. The severely affected regions are demarcated and twitter data is extracted using keywords respective to that location. The extraction of situational information from a large corpus of raw tweets adopts Content Word based Tweet Summarization (COWTS) technique. An integration of these modules using real-time location-based mapping and frequency analysis technique gathers multi-dimensional information in the advent of disaster occurrence such as the Kerala and Mississippi floods that were analyzed and validated as test cases. The novelty of this research lies in the application of segmented satellite images for disaster relief using highlighted land cover changes and integration of twitter data by mapping these region-specific filters for obtaining a complete overview of the disaster

# 1. Introduction

Disasters have a wide-ranging and long-term impact on nature and every community that constitutes the world. Some of the types of disasters include flood, earthquake, cyclone, volcanic eruption, wildfires etc. Disaster management is a broad term that encompasses all parts of emergency and disaster planning and response, including both pre and post-event operations [1]. Effective disaster management demands a sustainable infrastructure for the gathering, integrating and analyzing a diverse set of dispersed information sources, including real-time analysis via social media platforms like Twitter. In recent years, satellite image analysis techniques have developed to the extent that their effective application can tremendously assist in the management of natural disasters and humanitarian crisis circumstances. For a region-specific analysis, satellite images need to be outsourced from the affected locality to analyze the impact and magnitude of the disaster in study. The immensely damaged regions are further probed to extract more situational information via social media data. This integration allows for the catering of region-specific needs by providing immediate assistance and taking preferential rescue measures.

This research aims to analyse the pre and post disaster satellite images to elucidate the natural cover of the geographical region of interest. Further, the regional analysis is mapped to the requirements of basic amenities through social media and foster intensive priority-based rescue operations and offer support through proper planning. For instance, real-time emergency response scenarios such as flash floods calls for immediate planning of resource allocation and priority-based action. The case studies of Kerala and Mississippi floods were analyzed in two-steps, initially using satellite images where the most affected regions are localized through demarcation of land cover, followed by manually extracted social media data where the tweets from the specific regions are summarized with high priority for efficient response. Therefore, this research integrates different types of data (image and text) to obtain a complete understanding of disaster management analysis.

Section 2 discusses the existing work in natural disaster analysis and their shortcomings. Section 3 explains the proposed system in detail including the algorithms used. Section 4 discusses about the experimentation, results and performance analysis aspects. Section 5 is the concluding part that also elucidates the possible future work.

## 2. Related Works

The research on existing literature initially focuses on identifying the causes of disasters and gathering divergent perceptions specifically about climate change. We observe the trends in disaster damages related to life and economic losses and how effective early warning systems have proven to be in emergency-response situations. Furthermore, we explore empirical assessments that emphasize the way climate change translates to economic damages and the importance of including vulnerability and socio-economic factors for analysis. The data integration methods and the latest open-source technologies needed to build a consistent data corpus and visualizations were studied. Lastly, we examine suitable statistical approaches that could be used for analysis such as the correlation between factors and a comparative model for inferring results across various regions and types of disasters.

The existing literature on the applications of satellite image analysis in the field of disaster management focuses on categorizing the types of damages. Different approaches of feature extraction and classification are performed based on the type of satellite imagery including linear, aerial and UAV images.

Doshi et al. considers the image data from the Dak Nong province of Vietnam's senseFly UAV database to obtain disaster insights [2]. The input data consisted of 768 JPEG photos, which were collected and preprocessed using the OrthoEngine Tool in PCI in a 6-step procedure. The original UAV image was split into 12,000 sub-images and downsized to 128-pixel sub-images. As the outcomes of the prediction models are quantified values, IoU score, accuracy of class, overall accuracy, and Kappa coefficient were selected as evaluation metrics. Although this study discussed the scientific particulars and techniques involved in training a model for mining land covers, it is cumbersome to apply the same in real-time monitoring situations.

Tuan et al. proposed a framework for change detection on satellite images using Convolutional Neural Networks (CNN), which can then be thresholded and clustered together into grids to locate areas most severely affected by a disaster [3]. The framework achieves a top F1 score of 81.2% on the gridded flood dataset and 83.5% on the gridded fire dataset. As part of this work, they focus only on roads and buildings, however this can be extended to quantify disaster impact on other general natural and man-made features.

Alexander et al. performed the landcover classification task of the DeepGlobe Challenge using U-Net architecture and Lovasz-Softmax loss function which optimizes the Jaccard index to segment 7 classes of labels [4]. Chi et al. proposed a study that investigates land cover classification and change detection of urban areas from Very High Resolution (VHR) remote sensing images using deep learning-based methods [5].

Kavitha et al. proposed a method for generating the base map of a region in satellite imagery using efficient segmentation techniques [6]. The segmentation model was applied on multiple datasets collected from various sources each consisting of a particular land cover type namely, water bodies, vegetation, infrastructure/buildings and roads. Each segmented class is represented as a colored output which are then combined to create the segmented landcover base map of the region. The main architecture involved is U-Net along with ResNet-101 and VGG16 encoders as the backbone. The future scope of this study includes the application of land cover segmentation model for real-time disaster mitigation and damage estimation.

Twitter is a social media platform with over 1 million daily active users. This microblogging social network experiences a deluge of information flow during natural disasters [7]. The large volume and velocity of data flow on twitter during disasters makes it tedious for the disaster rescue volunteers to manually analyze and retrieve information from them. Capturing of twitter data can be done using Twitter API and authentication keys along with the python tweepy module. The proposed methodology uses event classification to prioritize tweets and extracts the address information for the high-priority tweets [8]. If the location cannot be inferred from the tweets, it uses Markov model to predict the location of the user from historical data. In context to the project aim, the location of the disaster is already fetched from the satellite image to be analyzed. Hence, a more comprehensive study is to be made regarding the accuracy of the requirements mentioned in the tweets pertaining to our use-case.

Koustav et. al proposed a methodology to extract and summarize situational information from twitter data during disasters [9]. The study considers Integer Linear Programming system for summarization of tweets related to Hyderabad bomb blast, Uttaranchal floods, Nepal Earthquake etc. The datasets with tweet ids are made available on the public forum which were outsourced for applying the summarization framework as it was the most suitable generic methodology for disaster specific scenarios.

The existing literature for the integration of social media data with satellite images comprises of methodologies for tracing most affected regions in a flooded area of the high-resolution input image [10]. Malika et al. proposed a system that uses Structural Similarity Index Measure (SSIM) difference to highlight demarcations based on the extent of damage in the post disaster satellite images. A Tensorflow object detection API is used to detect the presence of stranded people and map the coordinates in the images upload on social media platforms such as Twitter. The tweets are tokenized as single words, bigrams and trigrams to identify keywords of basic necessities required by people affected during disasters which are used in relief operations. The limitations of the research include the high dependence on intermittent network connection at the time of disaster occurrence, the scraping of a large proportion of unstructured data while identifying list of basic amenities which can be avoided by using query filters directly on the twitter platform.

Extensive research on related works in the remote sensing for disaster domain, the following observations are made: Existing Studies: The current research topics in the domain of disaster management have been confined to finding efficient methodologies to predict and identify the magnitude of the disaster and its regional impact. Another common research practice is dedicated to evaluating and gathering information post the disaster using satellite images to categorize different types of damages. Various machine learning algorithms for feature extraction and type classification have been implemented on different types of satellite images. Research gaps identified: The studies that implement land cover segmentation have limited its usage to geographical analysis with minimal types of covers being identified. The data collected for disaster management analysis is mostly confined to a single type, being satellite imagery or social media coverage, whereas an integration of useful filtered data from multiple sources is required to analyze the region-specific disaster situation. Proposed work with novelty: Owing to the research gaps noticed, we have developed a multi-class land cover segmentation model using U-Net architecture that can be analyzed for changes in geographical land cover according to the type of disaster that has occurred. For instance, floods show a significant increase in the water body cover, while earthquakes result in the demolition of buildings which can also be identified. Therefore, it is more feasible to collect different types of data related to natural disasters and apply suitable techniques using common attributes for integration to gather useful information for disaster management purposes. The novelty of the proposed research is implied in the real-time application of the land cover segmentation model and integration of social media data for outsourcing the region-wise impact to devise efficient relief strategies.

3. Methods

The proposed natural disaster analysis system considers two types of inputs such as the pre and post disaster satellite images and disaster related tweets and provides overall insight about the disaster situation as shown in Fig.1. The individual analysis from satellite imagery and social media data are combined for extracting region specific information about the particular disaster.

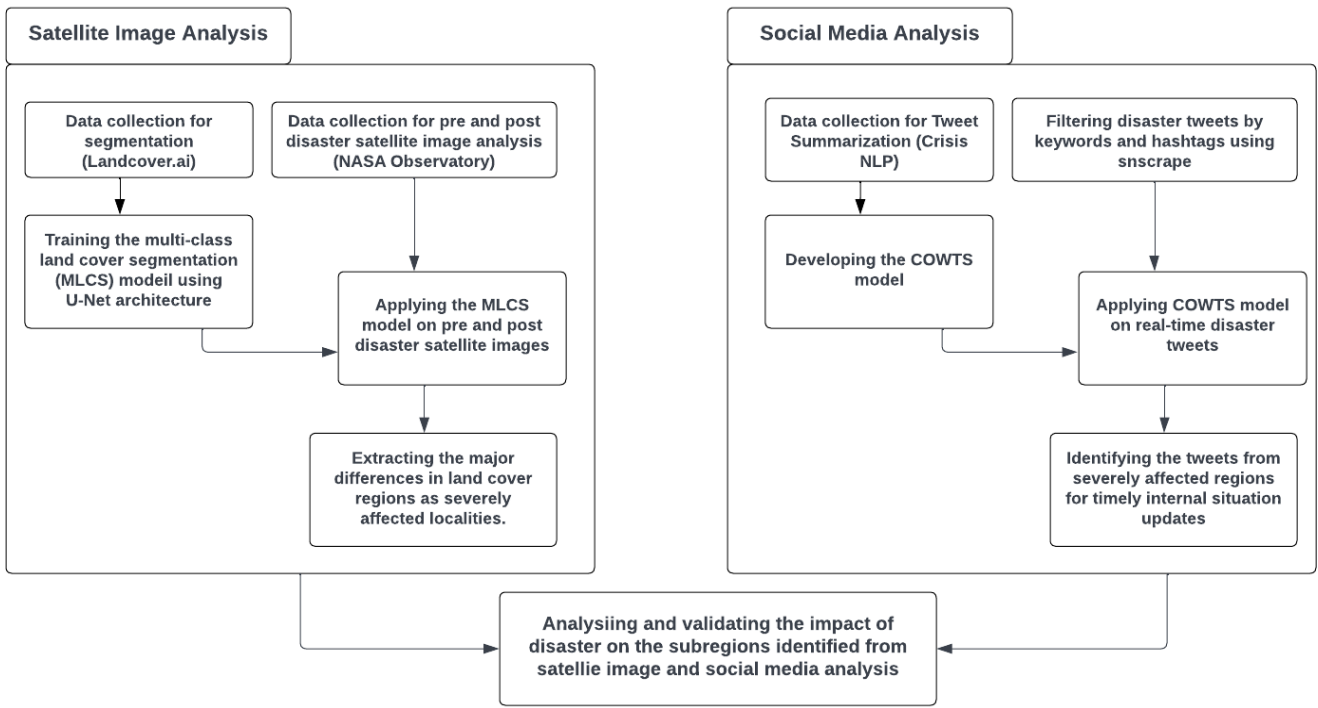

**Figure 1** Process Diagram

*3.1 Satellite Image Analysis*

The pre and post disaster satellite image analysis is implemented using U-Net based semantic segmentation and pixel-based feature extraction using machine learning concepts [11]. This section further elaborates in detail about the dataset collection, preprocessing and training aspects involved in the creation of multi-class segmentation model.

*3.1.1 Dataset Collection*

The satellite imagery data was obtained from LandCover.ai (Land Cover from Aerial Imagery) dataset for multi-class semantic segmentation [12]. A baseline analysis was done to check the performance of the LandCover.ai dataset and it stood out in terms of coverage and resolution with optimal number of classes. For applying the model on real-time satellite imagery, high resolution pre and post disaster images were outsourced from Nasa Earth Observatory [13]. Though there are multiple sources of image evidence of a disaster occurrence, it is essential that the obtained image for testing is of high resolution since it needs to be patched and tested by individual pixels.

*3.1.2 Dataset Preprocessing*

As part of processing, the following steps are executed sequentially –

(i) Read the 41 large images and corresponding masks, divide them into smaller patches of 256x256 and write the patches as images to the local drive.

(ii) Crop the images to a nearest size divisible by 256 and further divide all images into patches of 256x256x3 resulting in 41645 small patches.

(iii) Save only images and masks where masks have some decent number of labels other than zero since using blank images with label zero is a waste of resources and may bias the model towards unlabeled pixels.

(iv) Divide the sorted dataset from above into train and validation datasets, typically in the ratio 0.75: 0.25 respectively.

(v) Manually move some folders and rename appropriately to make use of the ImageDataGenerator module from keras.

*3.1.3 Multi-class Segmentation*

The U-Net architecture is applied for training the model which is available as a part of the segmentation models library. The ImageDataGenerator library enables the flowing of images and masks directly from the directory structure created in the data preparation phase. The multi-class segmentation model adopts an additive combination of categorical focal loss and Jaccard loss [14]. The model compilation uses Adam optimizer with IoU score metric during the training process.

A fundamental evaluation technique for multi-class segmentation is the mean Intersection over Union (mIoU). The IoU is formulated as the area of overlap between the actual pixels and the predicted pixels divided by their union [15]. Therefore, mIoU is defined as the average of intersection over union across all classes.

$$IOU = \frac{Intersection}{Union} = TP / (TP + FP + FN) \tag{1}$$

where 'TP' is True Positive, 'FP' is False Positive and 'FN' is False Negative.

Categorical Focal Loss: This loss function extrapolates multi-class SoftMax cross-entropy by incorporating the focusing parameter. This is suitable for segmentation since it increases the importance of correcting misclassified labels. Equation (2) defines categorical focal loss in a multiclass context with integer labels y.

$$L(y, \hat{p}) = -(1 - \widehat{p_y})^\gamma \log(\widehat{p_y}) \tag{2}$$

Where $y \in (0, \ldots, K-1)$ is an integer class label (K denotes the number of classes), and $\hat{p} = (\widehat{p_0}, \ldots, \widehat{p_{(k-1)}}) \in [0,1]^K$ is a vector representing an estimated probability distribution over the K classes.

Jaccard Loss: This function calculates the Jaccard index which is defined as the ratio between the overlap of the positive instances between two sets, and their mutual combined values is given by Equation (3)

$$J(A, B) = \frac{|A \cap B|}{|A \cup B|} = \frac{|A \cap B|}{|A| + |B| - |A \cup B|} \tag{3}$$

Jaccard loss is suitable for multi-class segmentation because of its perceptual quality and scale invariance, which lends appropriate relevance to small objects compared with per-pixel losses. Algorithm 1 shows how the loss, metrics and model are chosen and trained for the dataset.

The output of the training algorithm constitutes the accurate model which applied on patches that are smoothly blended to obtain a multi-class segmented image of the respective large satellite image. The steps for blending the image patches smoothly are given in Algorithm 2.

**Algorithm 1** Training the U-NET
___
**Input:** Dataset consisting of satellite images and its respective masks.
**Output:** Trained U-Net model
1: **function** TRAIN_MODEL(images,truth masks)
2:     Initialize model with the loss, accuracy metrics and the U-Net model
3:     Follow steps 4 to 8 for training the model
4:     **for** image in images **do**
5:         Predict mask for the image
6:         The loss and accuracy metrics are computed based on the predicted mask and ground truth mask with labels as given in Equations (1)-(3).
7:         Update model weights with respect to loss and accuracy metrics
8:     **end for**
9:     **return** *model*
10: **end function**
___

**Algorithm 2** Smooth Blending Image Patches
___
**Input:** The set of segmented 256x256 image patches
**Output:** Smooth prediction of large satellite image
1: **function** SMOOTH_WINDOWING_PREDICTION(input image, size of window, prediction function)
2:     Divide the 3D NumPy array image into patches by utilising 5D NumPy array for ordering
3:     **for** image in 256_patches **do**
4:         Reshape the images patches with an extra dimension of batch size to be used as a parameter in the prediction function of neural network
5:         Merge the patched predictions using the 5D NumPy array and combine spline interpolation to obtain 3D array of regular image
6:         Use rotations and mirroring to enable the neural network view of image in multiple angles
7:         Enlarge the input image and mirror-pad along its borders
8:         Apply the prediction function on each patch and merge them
9:     **end for**
10:     Unpad the resulting image to remove the extra borders predicted and save the final predicted large tile
11: **end function**
___

*3.2 Twitter Data Analysis*

This Section constitutes the methodologies involved in twitter data analysis including the dataset extraction, preprocessing and tokenization which is essential in formulating a concised summary of the disaster situation.

*3.2.1 Dataset Collection*

The tweets for implementing the summarization model are obtained from the dataset 'Twitter as a Lifeline: Human-annotated Twitter Corpora for NLP of Crisis-related Messages' [16]. This research categorized thousands of tweets from various disasters, such as the 2015 Nepal Earthquake, into different categories, such as 'displaced individuals and evacuations' and 'sympathy and emotional support'. The Twitter API allows for extraction of tweets using Tweepy only over the past seven days which has to be bypassed if a larger dataset of tweets has to be obtained. In order to fetch old tweets, snscrape module is used which is a scraper for social networking services (SNS) [17]. It scrapes things like user profiles, hashtags, or searches and returns the discovered items. Hence, it is used in order to extract real-time twitter data related to disasters as they are streamed on social media. The tweets related to Mississippi Flooding and Kerala Floods have been extracted to map the satellite image analysis of these regions with situational information.

*3.2.2 Dataset Preprocessing*

As per the privacy policy of Twitter, only tweet IDs can be saved to protect the information in case a tweet is deleted permanently. Therefore, Twitter API is required to retrieve the tweets by linking their respective identifiers from the dataset. The creation of a twitter developer account is necessary with elevated access. Furthermore, a project has to be created which provides authentication keys necessary for tweet extraction. The corresponding tweets for each tweet id in the corpus is fetched and a new corpus of tweets is generated for analysis. By filtering the valid tweet ids which still map to a particular tweet, 2779 tweets were extracted out of 3019 (about 92%). The data is loaded and empty tweets are not considered for analysis. For the tweets to be informative, few terms can be omitted. For instance, URLs and any '@...' which just calls another twitter handle and hashtags are removed.

*3.2.3 Tokenization and analysis of twitter data*

All the tweets extracted are not deemed to be useful for critical informative purposes. A significant proportion of non-situational tweets are present involving prayers, sentiments and opinions which do not offer much scope of analysis for emergency response. Whereas, situational tweets containing useful information such as status updates, seeking help or broadcasting helpline numbers are necessary for rescue operations. A Content-Word based Tweet Summarization (COWTS) model takes thousands of tweets as input and summarizes the ones that contain essential situational information related to the disaster. The necessary components and steps involved in developing the model are discussed below.

It is necessary to determine the characteristics using content words of the tweets that contribute to their efficacy. A technique for document analysis called term frequency-inverse document frequency (tf-idf) is applied to cluster the words which contain critical situational information in the tweets extracted regarding the occurring disaster. The SpaCy software is used to tokenize the tweets. SpaCy is a Natural Langauge Processing (NLP) package that analyses and retrieves textual information from a given document. It is an efficient method to identify content words. It combines extra attributes to the tokens, such as entity informaton, grammar tense, part of speech and sentimental category. It is presumed that highly valuable situational tweets are bound to have more content words when compared to non-situational tweets. The tf-idf score for some word t can be expressed mathematically using Equation (4).

$$tf - idf score = \overline{c}_t \cdot \log\left(\frac{N}{n_t}\right) \tag{4}$$

where c is the average number of times the word t appears in a document, N is the total number of documents, and n is the number of documents in which the word t appears.

Textacy is a tool that is added as an extension to SpaCy to evaluate the tf-idf scores of the words in the tweets. The total score of content words is maximized in the summary by defining some constraints using Equation (5) which is subjected to the following constraints.

$$\sum_{i=1}^{n} x_i + \sum_{j=1}^{m} Score(j) \cdot y_j \tag{5}$$

Where xi is 1 if tweet i is included, or 0 if not, yj is 1 or 0 if each content word is present and Score(j) is the tf-idf score of the word.

Constraint 1: The summary should not exceed a certain length and contain a definite number of words as represented by Equation (6).

$$\sum_{i=1}^{n} x_i \cdot Length(i) \leq L \tag{6}$$

The total length of all the selected tweets to be less than some value L, which is the length of summary.

Constraint 2: If a content word yj (out of m possible content words) is picked, then at least one tweet from the set of tweets which contain that content word, Tj has to be picked, as formulated in Equation (7).

$$\sum_{i \in T_j} x_i \geq y_j, j = [1, \dots, m] \tag{7}$$

Constraint 3: If a tweet i (out of my n possible tweets) is picked, then every content word in the tweet Ci is selected, as formulated in Equation (8).

$$\sum_{j \in C_i} y_j \leq |C_i| \times x_i, i = [1, \dots, n] \tag{8}$$

The defined constraints ensure a definite output with the scope of solving for variables in terms of chosen words and chosen tweets as defined in Equations (7) and (8). Using pymathproj library, this ILP problem is optimized to solve for x and y and obtain a summary of useful tweets as the output of the COWTS model.

## 4. Experiments and Results

### 4.1 Experimental Setup

The multi-class land cover segmentation model has been trained over the LandCover.ai dataset consisting of high resolution and simple RGB-only images. The dataset was formed through collecting and manually annotating images of 216.27 km^2 rural areas across Poland, 39.51 km^2 with resolution 50 cm per pixel and 176.76km^2 with resolution 25 cm per pixel. The dataset features satellite images with varying saturation, sunshine angles, display lengths and vegetative seasons. This strengthens and broadens the applicability of the dataset. Some sample images and their corresponding masks are shown in Fig.2 and the count of images in the dataset is summarized in Table 1.

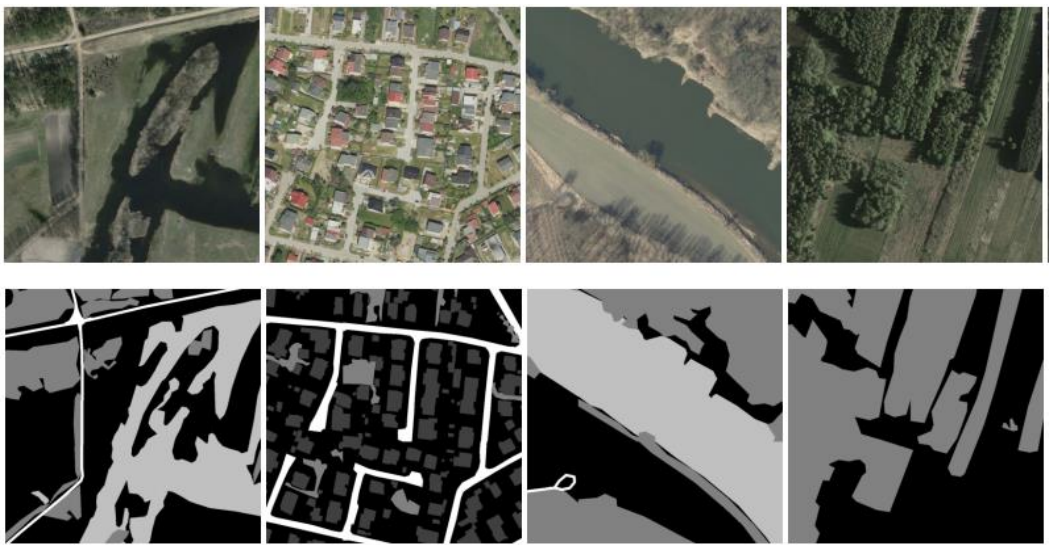

**Figure 2** Examples of satellite images and their respective masks.

| LandCover.ai Dataset | Image Count |
| --- | --- |
| Total Number of Images | 41645 |
| Number of Useful Images | 21570 |
| Number of Training Images | 15056 |
| Number of Testing Images | 5019 |

**Table 1:** Description of LandCover.ai Dataset

*4.1.2 Training and Validation*

The U-Net model was trained over 30 epochs as batches of 16 in the interest of time and resources with each epoch taking about 45 minutes runtime. After each epoch, a checkpoint was created with a saved model and the training variables were stored to load back and resume training from that point. The sample plot for the training and validation IoU score and loss at each epoch until

10 epochs is shown in Fig.3 and Fig.4

The patched images, masks and corresponding predictions are displayed in Fig.5 and Fig.6. The mean IoU score for a model of 30 epochs on a random batch of 16 images is 0.85.

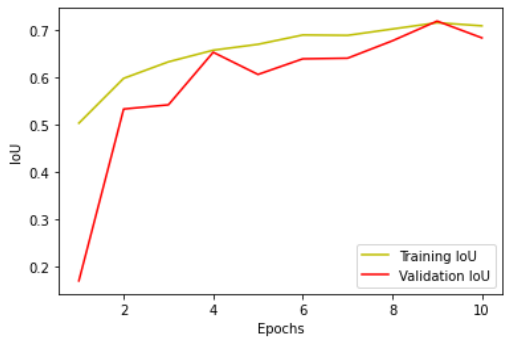 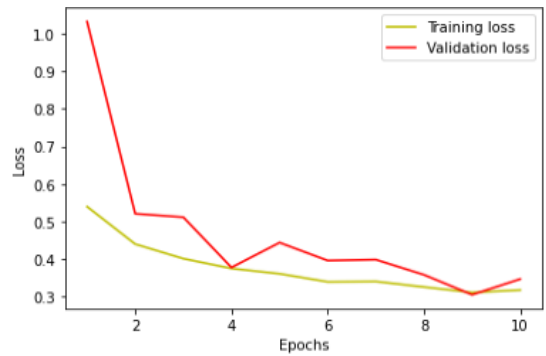

**Figure 3** Training and Validation IoU    **Figure 4** Training and Validation Loss

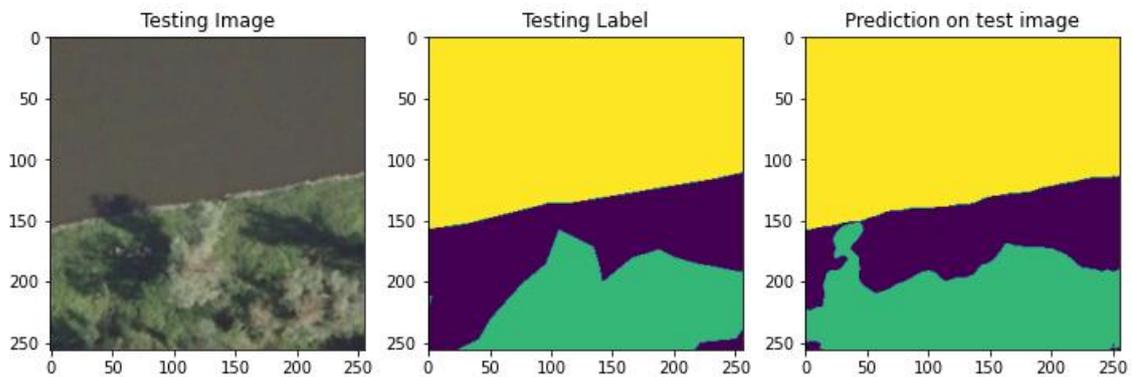

**Figure 5** Prediction of an image with water and vegetation.

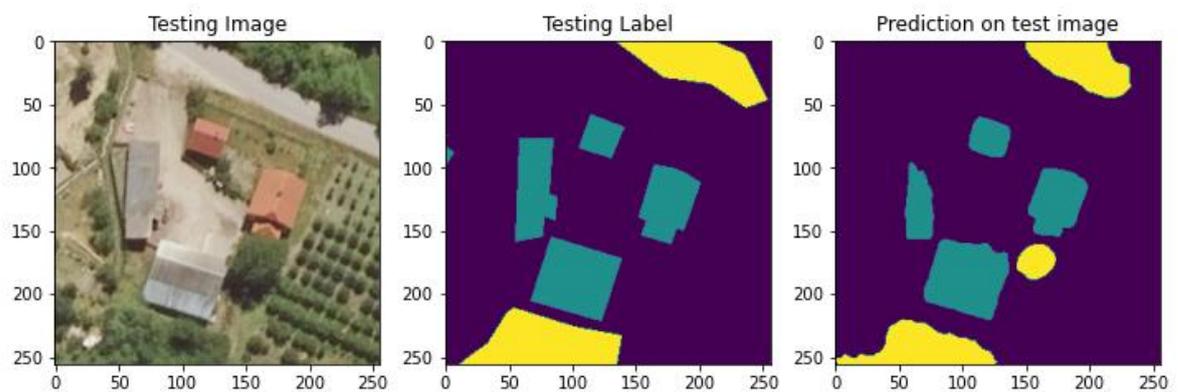

**Figure 6** Prediction of an image consisting of buildings and vegetation.

The predictions of 256x256 patches are then merged and blended smoothly to obtain the prediction of large image tiles as shown in Fig.7(a) - Fig.7(b).

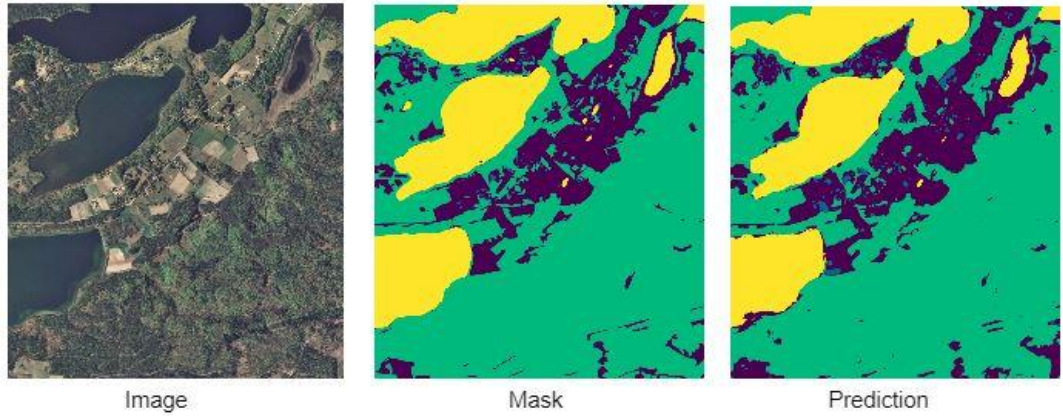

**Figure 7 (a)** Large image tile mask and corresponding prediction - 1

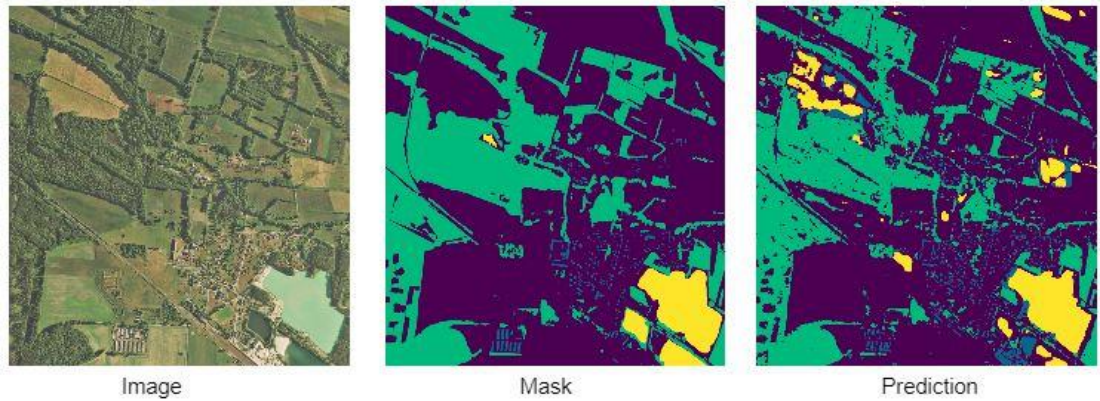

**Figure 7 (b)** Large image tile mask and corresponding prediction - 2

To realize the importance of patching the images and regrouping the segmented results, final predictions with and without best practices are compared in Fig.8

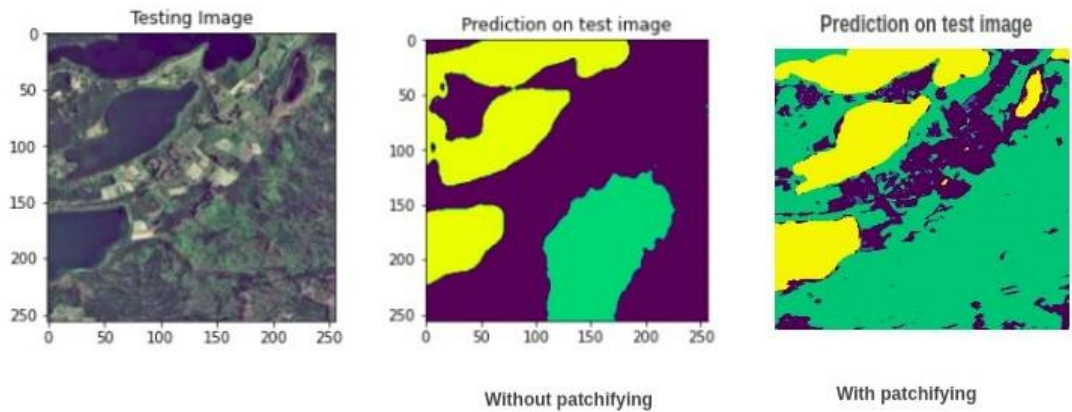

**Figure 8** Predicting an image without and with patchifying.

The multi-class segmentation results are narrowed down to the proportion of pixels classified correctly. The model classifies the pixels in one of the four labels, namely 0 for background class, 1 for buildings, 2 for vegetation and 3 for water class. The tabulated results of classification summarized in Table 2 suggest that the model performs the best in predicting background and vegetation classes accurately and fairly well in terms of predicting buildings and water geographical cover.

| Label | Predicted Pixel Count | Actual Pixel Count | Ratio |
| --- | --- | --- | --- |
| 0: Unlabeled Background | 458714 | 483062 | 0.9495 |
| 1: Buildings | 25805 | 28128 | 0.9714 |
| 2: Woodlands | 412782 | 416506 | 0.9910 |
| 3: Water | 70192 | 79692 | 0.8807 |

**Table 2** Model Performance Attributes

Once the segmentation model is obtained, it is applied on real-time satellite imagery of pre and post Missouri floods that are shown in Fig.9. and Fig.10.

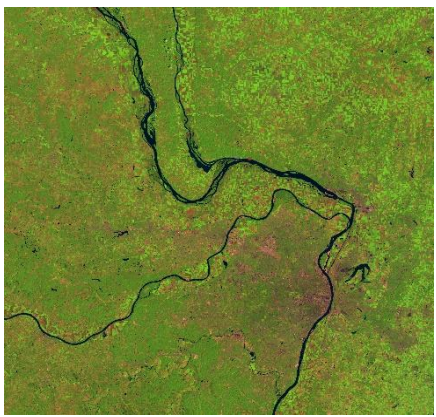

Figure 9 Satellite image of Missouri floods before disaster

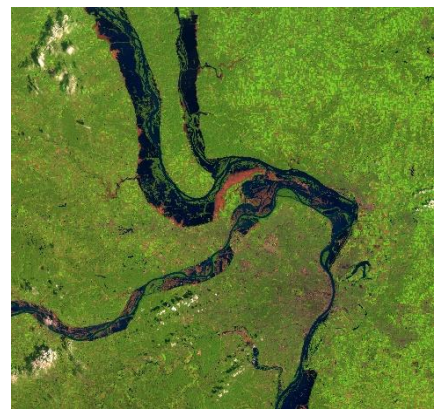

Figure 10 Satellite image of Missouri floods post disaster

During the first half of the year 1993, the great flood of the Mississippi River took place receiving more than 1.5 times their average rainfall [18]. In St. Louis, the Mississippi remained above flood stage for 144 days. The image pair 9 and 10 shows the area in August 1991 and 1993. After applying the segmentation model, in Fig.11 and Fig.12 we notice an increase in classified water pixels and a clear indication of the submerged vegetation in the river.

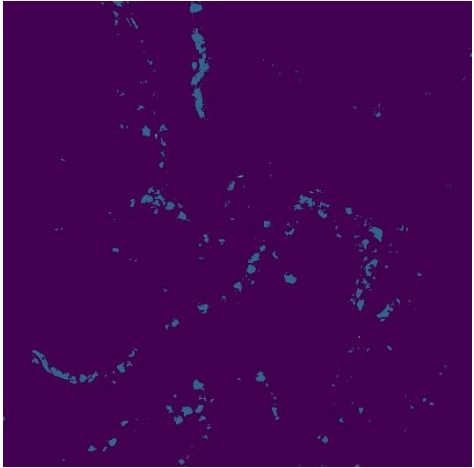 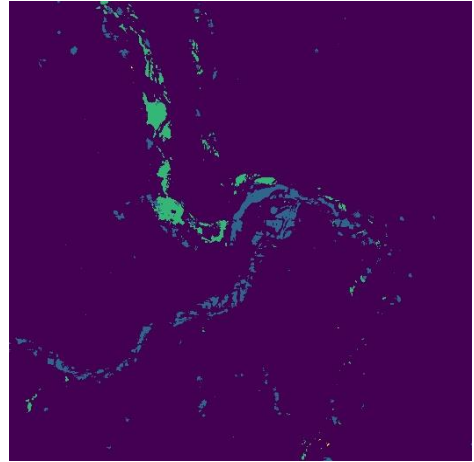

Figure 11 Segmented satellite image of Missouri before the flood

Figure 12 Segmented satellite image of Missouri after the flood

The application of the model on the real-time satellite image of the Mississippi river before and after the flood has resulted in location-wise. For the estimation of flood magnitude, we observe that the land-cover of the aquatic region has increased by 32.7652% and due to submergence of vegetation in the flood water, it is clearly depicted over water bodies, hence there is a multi-fold increase in the woodland region. The highlighted regions are correspondingly mapped as the most affected locations, namely along the rivers Mississippi, Missouri and Illinois. A few parts of North Dakota, Iowa and Kansas have also been severely affected.

A similar application of the model is carried out over the pre and post Kerala flood satellite images as shown in Fig.13 and Fig.14. The application of the model on the real-time satellite image of Kerala state before and after the flood enabled the identification of severely affected regions. A significant change in land cover was observed with a nearly 18.568% increase of water pixels. The differences in the pre and post flood satellite images highlighted Kochi, Alappuzha, Chengannur and Ambalapuzha as the most affected regions. In order to localize the most affected regions, the differences between the initial satellite image and the one captured after the disaster occurrence are highlighted as shown in Section 4.3 as a part of the performance analysis. The regions can be mapped using geographical annotations and further information regarding emergency response can be fetched using twitter data in real-time disaster scenarios.

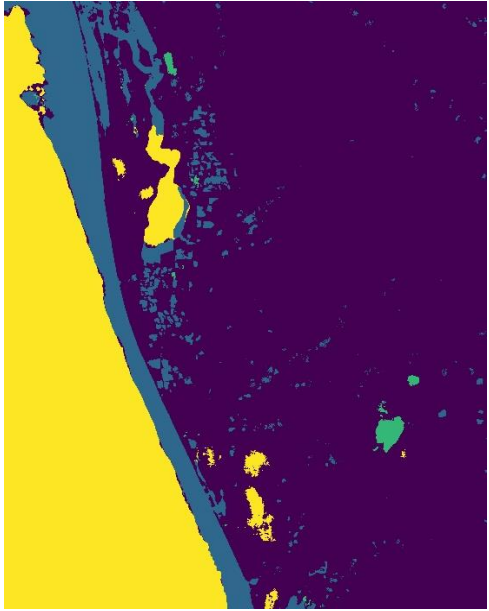 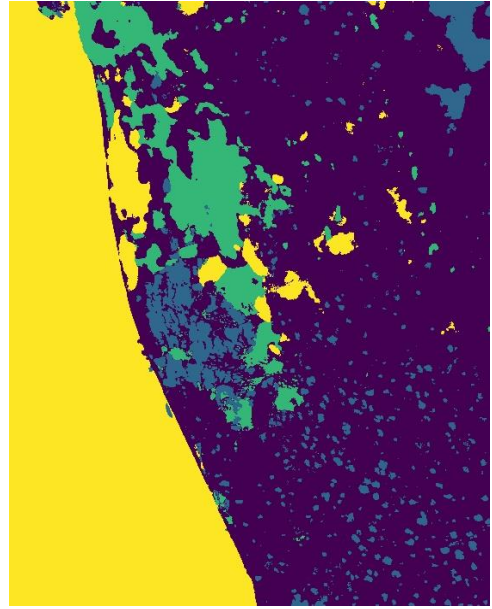

Figure 13 Segmented satellite image of Kerala before the flood

Figure 14 Segmented satellite image of Kerala after the flood

## 4.2 Twitter Data Analysis

### 4.2.1 Description of the dataset

The retrieval of useful consolidated information from tweets requires filtering through thousands of tweets and classifying them according to their importance in the scenario. Therefore, we obtained a dataset of thousands of disaster-related tweets from the corpus published in the paper 'Twitter as a Lifeline: Human-annotated Twitter Corpora for NLP of Crisis-related Messages' for validating the feasibility of extracting useful tweets that summarize the emergency situation. For the application of this summarization model on real-time streaming disaster data, tweets are scraped based on hashtags KeralaFloods and MississippiFloods using snscrape library. The sample data frame of the extracted real-time tweets is shown in Fig.15.

| Date | User | tweet_texts |
|---|---|---|
| 2019-12-21 16:42:18+00:00 | ArdraManasi | We are looking for pharmacists to volunteer to... |
| 2019-12-17 05:37:12+00:00 | pvrhere | #DamageToSabarimalaiTemple: Im very sorry that... |
| 2019-11-20 16:47:37+00:00 | jobinindia | #KeralaFloods2018 #KerlaFlood #KeralaFloodReli... |
| 2019-11-19 11:17:17+00:00 | Chanakyas_Rant | All @NCPspeaks MPs, MLAs and MLCs will donate ... |
| 2019-10-24 06:30:00+00:00 | indiawater | Thank you @timesofindia for taking a note. I a... |

Figure 15 Sample data-frame of scraped Kerala flood tweets with hashtag filters

*4.2.2 Twitter API Connection*

The extraction of tweets involves setting up a twitter developer account. As a next step, a project and an associated app needs to be created which will provide the following set of credentials that can be used to authenticate all requests to the API:

• API Key and Secret: Essentially the username and password for the App which will be used to authenticate requests that require OAuth 1.0a User Context or to generate other tokens.

• User Access Tokens: Represent the user that the request is made on behalf of.

By default, the access level is Essential which needs to be upgraded to Elevated access through an application that is reviewed and approved by the twitter team. This allows tweets to be extracted using Tweepy library over the past weeks' time. To overcome this limitation and fetch tweets from the full of twitter's archive, we can make use of the snscrape module to collect a wide range of tweets to build a corpus.

*4.2.3 Results of Twitter Data Analysis*

Using the snscrape tweet extraction methodology, numerous live tweets can be streamed at the time of disaster occurrence and summary of the most important tweets is presented for rescue operations. For instance, 5859 tweets during the Kerala flood of 2018 were scraped which consisted of noise and non-situational information. A list of main content words is created which are likely to involve relevant situational information. Some examples of such words are: 'rescue', 'food supplies', 'displaced', 'contact', 'urgent', 'help', contact numbers, internal regions of Kerala etc. The value of the content tokens is mapped to their tf-idf score using vectorizer vocabulary. The ILP methodology returns a set of chosen tweets that contain valuable situational information. The tweets were analysed and the situational summary obtained is presented in Fig.16.

```
Kerala, After The Flood: Water still to recede from many parts of state; people ask for clean clothes and sanitation
National Crisis Management Committee (NCMC) has directed that focus should now be on provision of emergency supplies and resto
ation of essential services as flood water recedes in Kerala. Catch all live updates here  KeralaFlood
In a bid to support livestock farmers, RelianceFoundation is conducting camps to provide invaluable medical treatment and rati
n for livestock. RFForKerala Keralafloods KeralaFloodRelief
1. Central assistance to flood affected Kerala: Centre has provided urgent aid and relief material in a timely manner and with
ut any reservation to the State. Situation has been regularly monitored by the PM on daily basis and he visited the State on A
gust 17-18 KeralaFloods
Appeal To: Stayfree India, Whisper Sanitary Napkin, Sofy Jockey India Carefree Liners Post-flood struggles of the displaced wo
en in Kerala: Relief camps in need of sanitary napkins
The rehabilitation efforts are also progressing in an admirable manner. Now there are 1,97,518 people from 53,703 families in
elief camps; which means 12,53,189 people (or 3,37,791 families) have returned to their homes. Rebuilding is now the task befo
e us. KeralaFloodRelief
@narendramodi Sir , Chenganoor area is in very dangerous condition..We need more army assistance there..Please Please help.@PM
India KeralaFloodRelief
50 k food packets required at Rajiv Gandhi stadium, Kadavanthra. Can anyone get bread packets from Modern or food packs and de
iver there? Please let us know. KeralaFloods
CM Pinarayi Vijayan reviewed the ongoing relief efforts. At present, there are 4,62,456 people in 1435 camps. The total death
ount from the rain havoc between August 8th and 26th is 302. KeralaFloods
Keralafloods | According to the preliminary assessment, standing crops cultivated on 56439.19 hectares are already destroyed.
he loss until August 21 was pegged at Rs 1,345 crore. reports
Kochi Indoor stadium filled with food, water and basic stationaries  please forward to all Pls contact +91 8593977444 KeralaFl
ods DoForKerala
```

Figure 16 Situational tweets summarization of Kerala floods, 2018

*4.3 Performance Analysis*

The performance analysis and validation of the integration is elucidated using the case studies of the Midwestern floods along the Mississippi River in USA in 2019, and the Kerala floods in 2018. The Midwestern United States experienced major floods in the spring of 2019 for an extensive time-period of nine months. The satellite image analysis highlights the most-affected regions using the pre and post disaster segmented results. The identification of most affected regions along the map shows severe flooding and vegetation submerging majorly along Nebraska and Iowa.

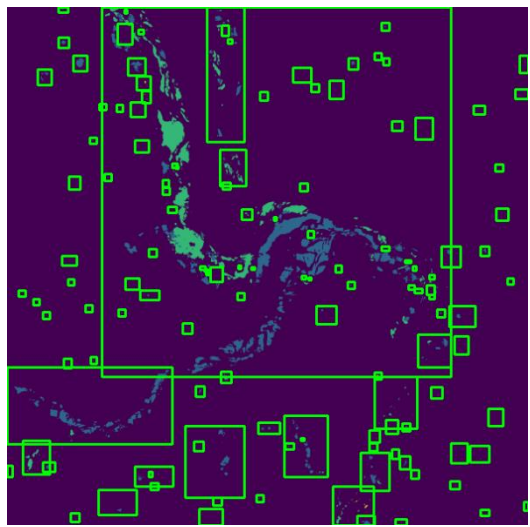

Figure 17 Highlighted regions changed from pre to post disaster satellite imagery.

A total of 1317 real-time tweets were extracted to obtain the dataset for text analysis. The proportion of tweets consisting of Nebraska is 75% whereas it is 37% for Iowa when compared to other

keywords and affected regions. It is also observed that most tweets related to Nebraska focus on vegetation and produce loss thereby validating the differences highlighted in post disaster segmented satellite image. Similarly, in the case of Kerala Floods, that occurred in 2018, the pre and post satellite images were segmented and their differences were highlighted as shown in Fig.18. The regions in the map where visual differences were obtained, were namely, Kochi, Alappuzha, Chengannur and Ambalapuzha.

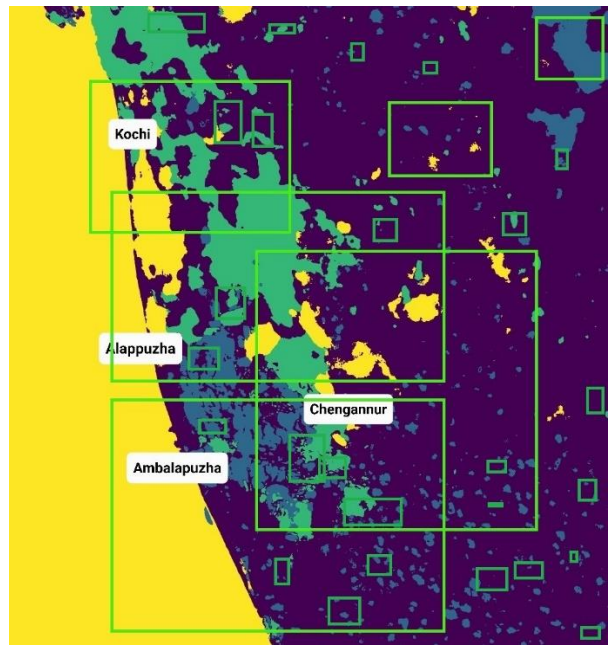

Figure 18 Most-affected regions highlighted in post Kerala floods satellite imagery

To validate this, a total of 5598 real-time tweets were extracted to analyze and develop a summary of the Kerala flood situation. The proportion of tweets consisting of information regarding Kochi were 56% followed by 32% for Chengannur, 28% related to Alappuzha and 17% related to Ambalapuzha and its surrounding regions, corresponding to the derived highlighted regions in the image analysis module.

The frequency of occurrence of keywords such as #floods, #resue, #keralafloods, #kochi, #alappuzha from the tweets fetched before the commencing of the disaster for a 10-day time period was 187. Owing to the spread of the natural disaster over the entire state, this count went up to 4739 in the subsequent time period of the same interval exhibiting nearly two thousand increases in percentage change. Similar analysis for the midwestern floods showcases that the tweets consisting of the keywords #mississippifloods, #nebraska, #food, #gdp were numbered at 989 in the previous year. However, this

scaled to 8264 in the subsequent year of 2019, resulting in a thousand-fold increase in reality

## 5. Conclusion and Discussion

The research aimed to integrate the analysis from satellite imagery and tweets extracted from social media. The satellite image analysis module is based on developing a multi-class land cover segmentation model using U-Net architecture with ResNet backbone. The dataset was obtained from LandCover.ai and the IoU evaluation metric is valued at 0.85 over 30 epochs. The model was applied on pre and post disaster satellite images of Kerala and Missouri and the corresponding severly affected regions were demarcated by highlighting changes between segmented images. The identified regions are mapped with social media analysis from twitter.

The dataset for developing COWTS model was outsourced from CrisisNLP and further tested on real-time tweets streamed during disaster occurrence. The snscrape library was used to bypass Twitter API limitations and obtain thousands of tweets for analysis. The COWTS model application is not confined to natural disaster tweets summarization. The scope of its usage extends to consolidating useful information about real-time happenings such as political, military and warfare situations. For instance, the Russo-Ukraine War of 2022 received immense social media attention, which is captured in the form of numerous real-time tweets that are similarly analyzed using the COWTS model.

The drawback of this study is that there is a limited availability of high resolution pre and post disaster satellite images. Therefore, only a handful of disasters could be considered for the integrated analysis. The machine learning techniques for super-resolution concept can be implemented to enhance the quality of low resolution pre and post disaster satellite images so they can be used for similar land-cover classification for various types of disaster analysis such as earthquakes, cyclones, drought and wildfires.